# Nonlinear variable selection with continuous outcome: a fully nonparametric incremental forward stagewise approach

Tianwei Yu[1]

[1] Department of Biostatistics and Bioinformatics, Rollins School of Public Health, Emory University, Atlanta, GA, USA. Email: tianwei.yu@emory.edu, Tel: (404) 727-7671

**Abstract**

We present a method of variable selection for the sparse generalized additive model. The method doesn't assume any specific functional form, and can select from a large number of candidates. It takes the form of incremental forward stagewise regression. Given no functional form is assumed, we devised an approach termed "roughening" to adjust the residuals in the iterations. In simulations, we show the new method is competitive against popular machine learning approaches. We also demonstrate its performance using some real datasets. The method is available as a part of the nlnet package on CRAN (https://cran.r-project.org/package=nlnet).

**Keywords:** variable selection, nonlinear association, forward stagewise regression

# 1. Introduction

Modern high-throughput biology and deep phenotyping data present the challenge of selecting a small subset of predictors from thousands of variables, which often exhibit complex correlation structure. Statistical variable selection for predictors linearly associated with the outcome variable has been extensively studied. Some major methods are reviewed in (Fan and Lv 2010, Wu and Ma 2014).

It is known that nonlinear and complex associations exist in omics data (Francesconi and Lehner 2014, Li 2002, Reshef et al. 2011). Such relations may represent critical regulatory mechanisms, and may be important for building robust predictive models. It is desirable to simultaneously select predictors that associate with the outcome variable both linearly and nonlinearly. Some existing regression and machine learning methods are aimed at achieving this goal. Regression methods include those that use polynomial approximation of nonlinear models (Rech, Terasvirta, and Tschernig 2001), functional and adaptive group Lasso (Huang, Horowitz, and Wei 2010, Ravikumar et al. 2009), regularization based on partial derivatives in reproducing kernel Hilbert space of smooth functions (Rosasco et al. 2010).

Testing procedures, such as the Hilbert-Schmidt Independence Criterion (HISC) (Gretton et al. 2005, Gretton et al. 2008), and Brownian Distance Covariance and its variations (Kosorok 2009, Székely and Rizzo 2009), can be used to detect nonlinear associations between random vectors. Such methods can be combined with some heuristic selection scheme to achieve variable selection. Examples include backward elimination using HSIC (BAHSIC) (Song et al. 2012), nonlinear independence screening (Wu et al. 2014), and independent ranking and screening (Zhu et al. 2011). Semi-parametric Copula regression can detect mild nonlinear relations (Noh, El Ghouch, and Bouezmarni 2013). However copula regression itself is not sensitive to non-monotone relations, and the success depends on the correct specification of the parametric family (Dette, Van Hecke, and Volgushev 2014). Lopez-Paz *et al* used random non-linear copula projections to overcome this issue, and achieved variable selection by greedy dependence maximization (Lopez-Paz, Hennig, and Sch¨olkopf 2013).

Some widely-used machine learning methods that provide variable importance ranking, such as Random Forest (Breiman 2001), multivariate adaptive regression splines (MARS) (Friedman 1991), and boosting (Friedman 2001) *etc*, are effective in selecting important predictors from large number of candidate variables. Given that nonlinear associations can be of different functional forms, and that high-throughput data generally contain higher levels of measurement noise, the statistical power to detect such associations and select the correct predictors is limited.

In this study, we consider the type of variable selection problem when the outcome variable is continuous, and it is associated with a small subset of *q* predictors in the form $E(Y|X_1, X_2, ..., X_q) = \sum_{i=1}^{q} f_i(X_i)$, where $f_1()$, $f_2()$, ..., $f_q()$ are arbitrary continuous functions. Variable selection under this sparse additive model setting has been explored by some authors, generally in the series expansions and regularized regression framework (Ravikumar et al. 2009, Huang, Horowitz, and Wei 2010). Here we present a variable selection method based on a fully nonparametric measure of nonlinear associations, and follow the general workflow of incremental forward stagewise regression, which can handle very large number of potential predictors (Hastie et al. 2007). Unlike the linear case, where forward stagewise regression can be achieved by gradually increasing the regression coefficients through the iterations, in our case there is no functional form assumed, and hence no regression coefficient. We devise a procedure named "roughening", which is the reverse of smoothing in concept, to allow the forward stagewise procedure in the nonlinear model-free scenario.

The nonparametric association method we use is the Dissimilarity based on Conditional Ordered List (DCOL) (Yu and Peng 2013, Yu, Peng, and Sun 2011), which is sensitive to relationships where an *X* has predictive value for *Y*, i.e. the distribution of *Y|X* is unimodal with limited spread. This is a useful property when our focus is selecting variables for prediction. In the following discussion, we use the abbreviation NVSD (Nonlinear Variable Selection using DCOL) to refer to our method. We demonstrate its performance in simulation studies, and its utility using real datasets.

## 2. Methods

*2.1. The model*

We assume the predictors form a p-dimensional random vector $(X_1, X_2, \ldots, X_p)$. Without loss of generality, we assume the first *q* variables truly associate with the continuous outcome variable *Y* through the functional form

$$E(Y|X_1, X_2, \ldots, X_q) = \sum_{i=1}^{q} f_i(X_i),$$

where $f_1()$, $f_2()$, ..., $f_q()$ are arbitrary continuous functions on finite support, with finite value and finite first derivative everywhere. We also assume all X variables are continuous and on finite support.

*2.2. Dissimilarity based on Conditional Ordered List (DCOL)*

We reported the DCOL and its utilities in missing value imputation and clustering in an *ad hoc* manner (Yu and Peng 2013, Yu, Peng, and Sun 2011). The DCOL is a useful measure of predictive nonlinear association between random variables/vectors (Fig. 1). Here we use a slightly different version of DCOL in order to estimate the variance component explained by each $X$ variable. Given two random variables $X$ and $Y$, and the corresponding data points $\{(x_i, y_i)\}_{i=1,\ldots,n}$, we sort the points based on the values of $x$ to obtain: $(x_i, y_i): x_1 \leq x_2 \leq \ldots \leq x_n$. We then obtain the $d_{col}(Y|X)$ by

$$d_{col}(Y|X) = 1/(n-1) \sum_{i=2}^{n}(y_i - y_{i+1})^2$$

Intuitively, when the spread of Y is small given X, $d_{col}(Y|X)$ is small (Fig. 1). Thus we can use $d_{col}(Y|X)$ to measure the spread of conditional distribution Y|X in a model-free manner.

*2.3 Estimating variance attributed to X in univariate regression without estimating the functional form between X and Y*

In this study we assume the relationship between $Y$ and $X$ is $Y = f(X) + \varepsilon$, where $f()$ is a continuous function, and $\varepsilon$ is additive noise with mean 0 and variance $\sigma^2$. We can use DCOL to estimate $\sigma^2$ without estimating the functional form of $f()$. Because

$$\Delta_i = y_{i+1} - y_i = f(x_{i+1}) - f(x_i) + \varepsilon_{i+1} - \varepsilon_i.$$

Assuming X is continuous on finite support, when the sample size is large, the difference between $x_{i+1}$ and $x_i$ approaches zero. Hence

$$\Delta_i \simeq \varepsilon_{i+1} - \varepsilon_i.$$

We also have:

$$\bar{\Delta} = \frac{1}{n}(f(x_n) - f(x_1) + \varepsilon_n - \varepsilon_1) \to 0, \text{ as } n \to \infty$$

Under the condition that $f()$ has finite value and finite first derivative everywhere, we can show (Appendix A) that

$$s(\Delta) = \frac{1}{n-2}\sum_{i=1}^{n-1}(\Delta_i - \bar{\Delta})^2 \to 2\sigma^2, \text{ as } n \to \infty \quad (1)$$

Thus if we take the sample variance of $\{\Delta_i\}_{i=1,\ldots,n-1}$, it provides an estimate of the variance of $\varepsilon$. Let

$$S_\Delta = \frac{1}{2(n-2)}\sum_{i=1}^{n-1}\Delta_i^2 = \frac{1}{2(n-2)}\sum_{i=1}^{n-1}(y_{i+1} - y_i)^2 \quad (2)$$

Then $S_\Delta$ is an estimate of $\sigma^2$. Given the sample variance of $\{y_i\}_{i=1,\ldots,n}$, $\hat{\sigma}_Y^2$, which is estimated directly from the sample, we have an estimate of how much of the variance of $Y$ is attributed to $X$, without knowing the function that links $Y$ to $X$.

### *2.4 Permutation test to assess the significance of Y's dependence on X.*

As the function linking $X$ to $Y$, $f()$, is unspecified, we can find the significance of the dependency of $Y$ on $X$ using the permutation test. Under the null hypothesis that $Y$ and $X$ are independent, sorting the data pairs $\{(x_i, y_i)\}$ based on the $x$ values is equivalent to a random re-ordering of $Y$. We repeatedly re-order the $\boldsymbol{y}$ vector in random to generate permuted vectors $\{\boldsymbol{y}^{(j)}\}_{j=1}^m$, and compute the $S$ value from the permuted vectors,

$$S_\Delta^{(j)} = \frac{1}{2(n-2)} \sum_{i=1}^{n-1} \left( y_{i+1}^{(j)} - y_i^{(j)} \right)^2, j = 1, \ldots, m.$$

We then take the proportion of $\{S_\Delta^{(j)}\}_{j=1}^m$ below the observed $S_\Delta$ to be the p-value of the permutation test.

The null distribution only depends on the $Y$ values, but not on the $X$ values. Thus no matter how many potential predictors we need to compare, the permutation only needs to be conducted on the $\boldsymbol{y}$ vector.

*2.5 Roughening*

The word "roughening" is used as opposed to "smoothing". We first describe the method, and then discuss its purpose in the next sub-section. As the name indicates, roughening is an anti-intuitive procedure that increase the roughness of the response of random variable $Y$ to a random variable $X$. In general, the procedure can be used with any smoother. Given the observations, $(x_i, y_i), i = 1, \ldots, n$, we can first fit any smoother to estimate the smoothed response at every given $x$, $(x_i, \tilde{y}_i), i = 1, \ldots, n$, and then calculate

$$\hat{y}_i = y_i + \theta(y_i - \tilde{y}_i), \qquad (3)$$

where θ is a small positive constant. This operation moves every point slightly away from the smoothed curve. Hence the name "roughening". The farther away the point is from its fitted value on the smooth curve, the more the point is moved. The size of the small constant may be heuristically determined. In this study, we used the cubic smoothing spline as the smoother, which fits a cubic spline, i.e. a piecewise cubic polynomial with continuous first and second derivatives, with a roughness penalty (Green and Silverman 1994).

Besides the general roughening procedure, we also develop a roughening process specifically for the DCOL. DCOL is not a smoothing procedure, yet its value is smaller when the relation between *Y* and *X* is smoother. Consider *eq. 2*, estimating a smooth curve would be to reduce the value of $S_\Delta$ with the fitted *Y* values. Thus the roughening procedure should go against the gradient to increase $S_\Delta$. Assume the *Y* values are ordered based on *X*,

$$\nabla S_\Delta = \frac{1}{(n-2)} \begin{pmatrix} y_1 - y_2 \\ 2y_2 - y_1 - y_3 \\ 2y_3 - y_2 - y_4 \\ \ldots \ldots \\ y_n - y_{n-1} \end{pmatrix}$$

With a small step size θ, which absorbs the constant term $\frac{1}{(n-2)}$, we go against the gradient to increase the value of $S_\Delta$.

$$y^{new} = y + \theta \nabla S_\Delta = \begin{pmatrix} y_1(1+\theta) - y_2\theta \\ y_2(1+2\theta) - (y_1 + y_3)\theta \\ y_3(1+2\theta) - (y_2 + y_4)\theta \\ \ldots \ldots \\ y_n(1+\theta) - y_{n-1}\theta \end{pmatrix} \qquad (4)$$

## 2.6 The effect of roughening

In an additive model with more than one predictors, by roughening based on one of the predictors, we change the relative contribution of the predictors, favoring other predictors. We start by discussing the DCOL roughening. We consider the situation where two predictive variables contribute to the generation of Y.

$$Y = f(X_1) + g(X_2) + \varepsilon$$

Without loss of generality, we can assume $f(X_1)$ and $g(X_2)$ both are centered at mean zero. In the roughening process, suppose we order the data points by $X_1$, such that

$$x_{1,i} \leq x_{1,i+1}, i = 1, \dots, N-1$$

At the same time, with the assumption that $X_1$ and $X_2$ are independent of each other, the ordering by $X_1$ has no bearing on $X_2$, i.e. $x_{2,i}, i = 1, \dots, N$ are i.i.d. samples from its underlying distribution. With DCOL roughening, we have

$y_i = f(x_{1,i}) + g(x_{2,i}) + \varepsilon_i$, and

$$y_i^{new}$$
$$= (1 + 2\theta)y_i - \theta y_{i-1} - \theta y_{i+1}$$
$$= y_i + \theta\left(2f(x_{1,i}) - f(x_{1,i-1}) - f(x_{1,i+1})\right) + 2\theta g(x_{2,i}) - \theta\left(g(x_{2,i-1}) + g(x_{2,i+1})\right)$$
$$+ \theta(2\varepsilon_i - \varepsilon_{i-1} - \varepsilon_{i+1})$$

Notice the data points are sorted based on $X_1$. Assuming all X's are on finite and continuous support, and f() is a continuous function, when the sample size is large, $x_{1,i-1}$ and $x_{1,i+1}$ both approach $x_{1,i}$. So $f(x_{1,i-1})$ and $f(x_{1,i+1})$ both tend to $f(x_{1,i})$. Thus the second term, which equals $\left(f(x_{1,i}) - f(x_{1,i-1})\right) + \left(f(x_{1,i}) - f(x_{1,i+1})\right)$, goes to zero.

On the other hand, given that the ordering has no bearing on $x_2$'s, thus $x_{2,i-1}$ and $x_{2,i+1}$ can be considered as random *i.i.d.* samples drawn from the probability density function of $X_2$. Hence $g(x_{2,i-1}) + g(x_{2,i+1})$ has a mean of zero, and variance of $2\varphi^2$, assuming the standard deviation of $g(X_2)$ is $\varphi$.

Thus we can argue that in $y_i^{new}$, the contribution from $g(x_{2,i})$ is on average boosted, as compared to $y_i$. We can write

$$y_i^{new} = y_i + 2\theta g(x_{2,i}) + \omega,$$

where $\omega$ is the noise term with mean 0 and variance of $2\theta^2\varphi^2 + 6\theta^2\sigma^2$. Hence after a step of roughening, the relative contribution of the predictive variable that is not the basis of the current roughening step would be increased. The same argument can extend to later iterations of the roughening process, as well as the scenario of multiple predictive variables. When the general roughening procedure is used with a smoother, a set of positive weights are applied to the points surrounding the $i^{th}$ data point. The above argument still holds. See Appendix B for details. Notice the arguments here require the assumption of $X_1$ and $X_2$ being independent to be true.

## 2.7 Incremental forward stage-wise variable selection procedure

In the linear regression framework, regularized regression provides an effective approach to select predictors from a large number of variables. However in the nonlinear framework and without any assumption on the function linking the outcome to the predictors, regularization cannot be easily achieved. It has been shown that forward stagewise regression achieves similar effect as $L_1$ regularization in linear regression (Hastie et al. 2007). Here we devise a forward stagewise regression procedure for nonlinear regression.

In each step of the forward stagewise selection, our goal is to take out a small portion of the contribution by the currently selected variable. When linearity is assumed, this is easily done by conducting linear regression and adding gradually and iteratively to the regression coefficients. Alternatively, the new residual can also be obtained by adding "noise" with regard to the currently selected variable $x^*$ to the residual. The idea behind the procedure is that the "error" with regard to the current predictor $x^*$ contains true signal from other predictors, as argued in the previous section. Here we do not wish to assume a functional form. So instead we use the roughening procedure to add errors with regard to $x^*$ to the residual vector.

In deciding which predictor is best associated with the current residual in every iteration, we consider the fact that when the true underlying relation is linear, Pearson's correlation has higher statistical power than non-linear association methods. Thus we take a heuristic approach: we compute both the Pearson correlation and its p-values, and the DCOL-based p-value. Then the minimum of the two p-values is taken, and multiplied by 2 for a simple Bonferroni-type correction. Box 1 shows the workflow of our procedure.

In every iteration, the top-ranked candidate predictor is selected for the roughening procedure. In the case where the predictors are mutually independent, it is easy to see that when the sample size is large, all true predictors will receive higher rank than nuisance predictors, and based on our discussion in the previous section, when the sample size is large, the roughening process doesn't improve the ranks of nuisance predictors, but changes the rank within the true predictors.

**Box 1. Forward stagewise variable selection based on DCOL**

(1) Set θ to some small constant, such as 0.01.

(2) Find the p-values of linear association and DCOL association between every $X$ and $Y$, $p_i^{(linear)}$ and $p_i^{(DCOL)}$, $i = 1, \ldots, p$.

For every predictor $X_i$, take $p_i = 2min\left(p_i^{(linear)}, p_i^{(DCOL)}\right)$.

(3) Find the predictor with smallest p-value, and conduct the roughening procedure with step size θ.

(4) With the updated Y values, repeat steps (2) and (3).

(5) Stop the iteration until the minimum p-value is larger than a predetermined threshold, such as 0.001.

Given no functional form is assumed, our procedure doesn't include a prediction model once the variables are selected. Existing nonlinear regression models can be borrowed to make predictions with the selected variables. In this study we used the multivariate adaptive regression splines (MARS) model (Friedman 1991) for prediction.

## 3. Results and Discussion

*3.1 Simulation study*

We conducted a simulation study using the following data generation scheme:

(1) Determine the number of true predictors $q$, the total number of potential predictors $p$, the sample size $n$.

(2) Generate the matrix $X_{p \times n}$ of observations for all potential predictors. Introduce correlation between the predictors by generating multivariate normal data using the correlation structure of $p$ randomly sampled genes from a real gene expression matrix (Spellman et al. 1998). When uniformly distributed predictors are needed, each row of the data matrix is transformed by taking normal quantiles.

(3) Randomly select $q$ rows of the matrix to be true predictors. For the $i^{th}$ true predictor, randomly draw a function $f_i()$ that links it to the outcome variable: linear (50% chance), absolute value (12.5% chance), sine (12.5% chance), sawtooth wave (12.5% chance), and box wave (12.5% chance). Randomly draw a coefficient from unif[1, 3], and with 50% chance flip the sign of the coefficient.

(4) Generate the $y$ values by $y_j = \sum_{i=1}^{q} \beta_i f_i(x_{ij}) + \varepsilon_j$, with $\varepsilon_j$ are *i.i.d.* samples from $N(0, \sigma^2)$.

After data generation, we split the data into the training and testing data at a 1:1 ratio. The training data was analyzed by four different methods: NVSD with cubic smoothing spline roughening and DCOL roughening, generalized boosted regression with Gaussian (squared error) loss (Friedman 2001), Random Forest (RF) for continuous outcome (Breiman 2001), multivariate adaptive regression splines (MARS), and backward elimination using HSIC (BAHSIC) (Song et al. 2012). Using the training data, each method was used to rank the candidate predictors, and a 5-fold cross-validation was conducted sequentially from the most important variable to determine the best number of predictors. Then prediction was conducted on the testing data. For BAHSIC, seven kernel settings were tested, which include linear kernel, inverse distance kernel, and Gauss kernel with scale parameter 1000, 10, 1, 0.1 and 0.001. Prediction was conducted using MARS. The best-performing among the five in each simulation scenario is reported. Given the high correlations between the candidate predictors, and each method's different level of resistance to nuisance variables, we decided to use prediction accuracy on testing data to compare the performance of the methods. The prediction accuracy was evaluated using a modified version of normalized root mean squared error (NRMSE):

$$NRMSE = \sqrt{\sum_j (\hat{y}_j - y_j)^2} / range(y).$$

For all the methods, cross-validation was used to select the number of predictors, and the prediction accuracy was found using only the selected predictors.

We used a number of parameter combinations, *i.e.* sample size, number of true predictors, total number of variables, with each setting repeated 50 times. Figure 2 shows the results of average NRMSE. The rows represent different numbers of candidate predictors, and the columns represent different numbers of true predictors. NVSD with cubic smoothing spline roughening (solid red line) showed slightly better performance than NVSD with DCOL roughening (dashed red line) in most scenarios. When the true number of predictors is 3, NVSD clearly out-performed the other methods at low to moderate sample sizes (Fig.2, left column). When the true number of predictors is 6, NVSD performed similarly to BAHSIC, boosting and RF at low sample size, while maintaining an edge at moderate sample sizes.

When the sample size was large, NVSD fell slightly behind MARS (Fig.2, center columns). When the true number of predictors was increased to 15, Boosting and RF achieved better performance at low sample size, and NVSD remains competitive at moderate to high sample sizes (Fig.2, right column). When the number of nuisance variables is small, BAHSIC achieved the best performance when sample size was large. When the number of nuisance variables is large, boosting performed better when the sample size was large.

Overall, NVSD behaved quite competitively against the other popular machine learning methods, especially when the sample size is not large. Besides uniformly distributed predictors, the results generated from normally distributed predictors are shown in Supplementary Figures 1, which generally agrees with Figure 2. Although NVSD was designed for continuous outcome data, we tested its performance on data with binary outcomes. The data generation followed the same procedure as described. After the y values were generated, we further dichotomized the values into two groups by thresholding at the median. NVSD was again compared with the four other methods using prediction error rate as the performance indicator (Supplementary Figure 2). Similar to the continuous case, NVSD had an edge when the true number of predictors was relatively small, and remained competitive when the true number of predictors became larger.

The roughening procedure iterates until no predictor is significantly associated with the roughened residuals. The step size parameter $\theta$ controls the rate of change of the residuals in the roughening procedure. In the current results we used $\theta=0.005$. In Supplementary Figures 3 and 4, we show comparisons of $\theta=0.001$, $0.005$, $0.01$, and $0.05$. They were applied on the same data, with 10 independent datasets at each parameter setting. The results were almost identical, except the smallest step size 0.001 performed slightly worse than the rest in a few settings using spline. The results indicate the method is robust against the choice of θ in a reasonable range.

The correlation structure between the predictors may influence the performance of the methods. To assess the impact, we conducted three more simulations – (1) The covariance matrix is the identity matrix (Supplementary Figure 5); (2) weaker correlation matrix - Consider the covariance matrix is $\Sigma$ from the real data. In the modified matrix $\Sigma'$, $\sigma'_{ij} = sign(\sigma_{ij})\sigma_{ij}^2$ (Supplementary Figure 6); (3) a stronger correlation matrix - Consider the covariance matrix is $\Sigma$ from the real data. In the modified matrix $\Sigma'$, $\sigma'_{ij} = sign(\sigma_{ij})\sqrt{|\sigma_{ij}|}$ (Supplementary Figure 7). In all three situations, the relative performance between the methods generally stayed the same.

*3.2 Community crime rate data*

The Communities and Crime Data Set was downloaded from the UCI machine learning data repository (Lichman 2013). The data contains 1994 communities (rows) and 123 attributes (columns) (Redmond and Baveja 2002). The outcome variable is community crime rate. Some missing values were present. After removing attributes with >10% missing values, 90 attributes were retained for the analysis. K-nearest neighbor (KNN) imputation was used to impute the remaining missing values.

We applied the forward stagewise variable selection procedure to the data, with a stopping alpha level of 0.01. Eight variables were selected by this procedure (Table 1). Three of the variables (PctKids2Par, FemalePctDiv, PctIlleg) are related to family structure; two variables (racePctWhite, racePctHisp) are related to race; three variables (pctWInvInc, PctPersDenseHous, HousVacant) are related to the housing conditions of the region.

As shown in Figure 3, six of the selected variables, racePctWhite, PctPersDenseHous, HousVacant, PctIlleg, HousVacant, and racePctHisp, showed clear nonlinear relations with the outcome variable. We then obtained predicted values from 5-fold cross-validation using the same variable selection procedure (Figure 3, lower-right panel). For $y$ values at the lower end (low crime rates), the prediction appears to be biased towards higher values. Otherwise the prediction is reasonably good.

All methods used in the simulation were applied to the data and prediction accuracy was compared using NRMSE. The data was randomly split into training and testing datasets at a 1:1 ratio for 20 times. For BAHSIC the kernel was selected using 5-fold cross validation in the training data. And the average NRMSE on the testing data was calculated. Overall the performance were close, with NVSD and gbm leading the performance at NRMSE=0.136. The NRMSE of other methods were MARS: 0.137; RF: 0.141, and BAHSIC: 0.140.

*3.3. Gene expression in ALL patients (GSE10255)*

We downloaded the GSE10255 dataset from the Gene Expression Omnibus (GEO) (Barrett and Edgar 2006). The data contained gene expression in diagnostic bone marrow leukemia cells in patients with primary acute lymphoblastic leukemia (ALL). The dataset is measured with HG-U133A gene expression microarray. We selected the probesets with known ENTREZ Gene IDs.

For genes represented by more than one probesets, we merged the corresponding probesets by taking their mean expression levels. The dataset contained 12704 genes and 161 samples. The outcome variable is the reduction of circulating leukemia cells after MTX treatment. Here the interest is mainly in selecting genes that are relevant to the disease outcome and analyzing the biological implications of such genes.

Given the dataset contains magnitudes more genes than samples, we used an iterative procedure to select multiple groups of genes. This is a heuristic approach necessitated by the fact that the biological system is modular - genes function in quasi-autonomous groups (Wagner, Pavlicev, and Cheverud 2007), and each of the groups may respond almost independently to the clinical situation (Ideker and Krogan 2012). We expect that multiple biological functions (gene modules) may be associated with the clinical outcome, and each module may predict the outcome well by itself `given the limited sample size.`

We first conducted the NVSD to select a group of genes. Then after removing the selected genes from the data matrix, we applied the NVSD again to select another group of genes. This process was iterated until the group size was less than 20. A total of 17 groups were selected. The full list of genes are in Appendix D. The first one contained 134 genes. We used the GOstats method to evaluate the biological functions of each group (Falcon and Gentleman 2007), based on the Gene Ontology biological processes. We limited the analysis to GO biological process terms with 10 to 1000 human genes. We show the top 5 GO terms of the first 5 groups in Table 2.

The first group (134 genes) over-represents some signal transduction pathways, including the granulocyte macrophage colony-stimulating factor (GM-CSF) production, the JAK-STAT cascade, as well as immune cell proliferation. Given that MTX is an immune suppressor, it is expected that immune cell proliferation processes are related to the MTX treatment outcome. At the same time, the JAK-STAT pathway has been documented to be related to the disease ALL. Mutations in JAK1 and JAK2 can cause constitutive JAK-STAT activation, which is associated with ALL (Mullighan et al. 2009, Hornakova et al. 2009). On the other hand, it was suggested that constitutive JAK-STAT activation could also be achieved through an autocrine loop involving GM-CSF (Chai, Nichols, and Rothman 1997). Similarly, groups 2 and 3 also over-represents some immune, stress response, and signal transduction GO terms. Group 4 over-represents GO terms of cell motility and regulation in protein degradation. Some genes involved in these processes and selected by the NVSD method have been documented to be important in leukemia. For example, TRIB1 was found to be important in myeloid cell development and transformation (Nakamura 2015), and APOE was found to be an important marker in distinguishing high- and low-risk pediatric ALL (Braoudaki et al. 2013). Among the terms over-represented by group 5, plasminogen activation was documented in some acute leukemia cells, and thought to contribute to the invasive behavior of these cells (Scherrer et al. 1999).

We next examined the first group of genes in more detail. Among the 134 genes, 47 have very low linear correlation with the outcome variable. The absolute value of Spearman's correlation coefficients between these genes and the outcome are below 0.1. We further examined the biological functions over-represented by this subset of 47 genes. As shown in Table 3, the top 5 GO terms were still dominated by GM-CSF signal transduction and some immune system processes, including cytokine and immune cell proliferation terms. The results agree well with Table 2, which indicate that the variables found by the NVSD method were not dominated by those linearly associated with the outcome variable, and those variables nonlinearly associated with the outcome are functionally meaningful.

*3.4. Discussion*

In this study, we devised a nonlinear variable selection scheme for continuous outcome named NVSD (Nonlinear Variable Selection using DCOL). It is a nonparametric incremental forward stagewise procedure. No functional form between the predictors and the outcome variable is assumed. The implementation of the method is available as part of the nlnet package on CRAN (https://cran.r-project.org/package=nlnet), which contains a number of methods based on DCOL. The computation time is shown in Supplementary Figure 8. The figure was generated using step size of 0.01, stopping alpha of 0.01, and in the setting of 6 true predictors, on an iMac computer with Core i7-860 CPU. Using different number of predictors resulted in similar computing time. The DCOL version and the spline version used similar computing time. Hence we report the average computing time over the two in the figure. Empirically, the computing time increased roughly linearly with the sample size, and roughly quadratically with the number of predictors. With 100 predictors and 100 samples, the method took ~0.5 minute. With 1000 predictors and 500 samples, the method took ~8 minutes.

Although the NVSD method assumes no functional form, hence no coefficient is available, we implemented a heuristic approach to show the solution path. Based on section 2.3, suppose variable $X_i$ is first selected at step $k$, we can estimate the proportion of variance of the residual $r_k$ that is attributed to $X_i$, denoted $S_{\Delta,k}$. Then at step $j$, if $X_i$ is selected again, we can estimate the proportion of variance of the current residual $r_j$ attributed to $X_i$, denoted $S_{\Delta,j}$. We then take the ratio $d_j=(S_{\Delta,k}- S_{\Delta,j})/S_{\Delta,k}$. It can be easily seen that in the simple case of a single predictor being repeatedly selected, $S_{\Delta,j}$ decreases with the roughening steps, as the predictor's contribution to the residuals becomes smaller and smaller. Thus $d_j$ increases with the iterations and approaches

1. With multiple predictors, the relation becomes more complex, and $d_j$ only roughly records the reduction of the proportion of the outcome being explained by $X_i$ along the iterations. A plot of $d_1, ...., d_m$ against the step number *1, ..., m*, where *m* is the total iterations, resembles the coefficient path plot of the linear forward stagewise regression, but without the relative scale between the predictors. An example plot derived from the crime rate data is shown in Figure 4.

There is an conceptual relation between NVSD and boosting. We draw a parallel to the linear case. For linear regression, the incremental forward stagewise regression can be seen as a version of boosting, achieved by a subgradient descent to minimize the correlation between the residuals and the predictors (Freund, Grigas, and Mazumder 2013). In the NVSD using DCOL roughening, we are indeed conducting a gradient descent of the relation as measured by the DCOL statistic. In the NVSD using smoothers, points farther away from the smoothed curve are moved by a larger amount in the generation of the new residuals. Although motivated from a different angle, it is conceptually similar to boosting with $L_2$ loss and component-wise smoothing spline as the learner (Bu˙hlmann and Yu 2003). We cannot call NVSD a boosting procedure, because it is not directly aimed at minimizing a loss function for prediction, but we see it is connected to boosting in concept.

**Acknowledgements**

This work was partially supported by NIH grants R01GM124061 and U19AI090023.

**Tables and Figures**

**Table 1. Selected variables for the communities crime rate data.**

| Variable | Attribute |
|---|---|
| PctKids2Par | percentage of kids in family housing with two parents |
| racePctWhite | percentage of population that is Caucasian |
| FemalePctDiv | percentage of females who are divorced |
| pctWInvInc | percentage of households with investment / rent income in 1989 |
| PctPersDenseHous | percent of persons in dense housing (more than 1 person per room) |
| PctIlleg | percentage of kids born to never married |
| HousVacant | number of vacant households |
| racePctHisp | percentage of population that is of hispanic heritage |

**Table 2.** Top 5 GO biological process terms for the first 5 groups of genes.

| GOBPID | Pvalue | Term |
|---|---|---|
| Group: 1 , number of genes: 134 | | |
| GO:0032645 | 0.000161 | regulation of granulocyte macrophage colony-stimulating factor production |
| GO:0046427 | 0.00053 | positive regulation of JAK-STAT cascade |
| GO:0046641 | 0.00086 | positive regulation of alpha-beta T cell proliferation |
| GO:0032946 | 0.00104 | positive regulation of mononuclear cell proliferation |
| GO:0050714 | 0.0014 | positive regulation of protein secretion |
| Group: 2 , number of genes: 120 | | |
| GO:1901998 | 0.0028 | toxin transport |
| GO:2001240 | 0.0028 | negative regulation of extrinsic apoptotic signaling pathway in absence of ligand |
| GO:0034405 | 0.00308 | response to fluid shear stress |
| GO:0045619 | 0.0036 | regulation of lymphocyte differentiation |
| GO:0048566 | 0.00401 | embryonic digestive tract development |
| Group: 3 , number of genes: 107 | | |
| GO:0002819 | 2.55E-06 | regulation of adaptive immune response |
| GO:0050707 | 1.76E-03 | regulation of cytokine secretion |
| GO:0051223 | 2.69E-03 | regulation of protein transport |
| GO:0035058 | 3.22E-03 | nonmotile primary cilium assembly |
| GO:0007168 | 3.91E-03 | receptor guanylyl cyclase signaling pathway |
| Group: 4 , number of genes: 60 | | |

| GO:2000146 | 0.000125 | negative regulation of cell motility |
| GO:0051928 | 0.000596 | positive regulation of calcium ion transport |
| GO:0043903 | 0.00147 | regulation of symbiosis, encompassing mutualism through parasitism |
| GO:0045862 | 0.00151 | positive regulation of proteolysis |
| GO:0043243 | 0.00333 | positive regulation of protein complex disassembly |
| Group: 5 , number of genes: 62 | | |
| GO:0010755 | 0.000934 | regulation of plasminogen activation |
| GO:0002407 | 0.0031 | dendritic cell chemotaxis |
| GO:1901522 | 0.00422 | positive regulation of transcription from RNA polymerase II promoter involved in cellular response to chemical stimulus |
| GO:0048384 | 0.00964 | retinoic acid receptor signaling pathway |

**Table 3.** Top 5 GO biological process terms for the genes in group 1 that are not linearly correlated with the outcome variable (absolute value of Spearman correlation less than 0.1).

| GOBPID | Pvalue | Term |
| --- | --- | --- |
| GO:0032725 | 6.79E-06 | positive regulation of granulocyte macrophage colony-stimulating factor production |
| GO:0042119 | 6.27E-05 | neutrophil activation |
| GO:0050730 | 0.000513 | regulation of peptidyl-tyrosine phosphorylation |
| GO:0042108 | 0.00111 | positive regulation of cytokine biosynthetic process |
| GO:0046634 | 0.00138 | regulation of alpha-beta T cell activation |

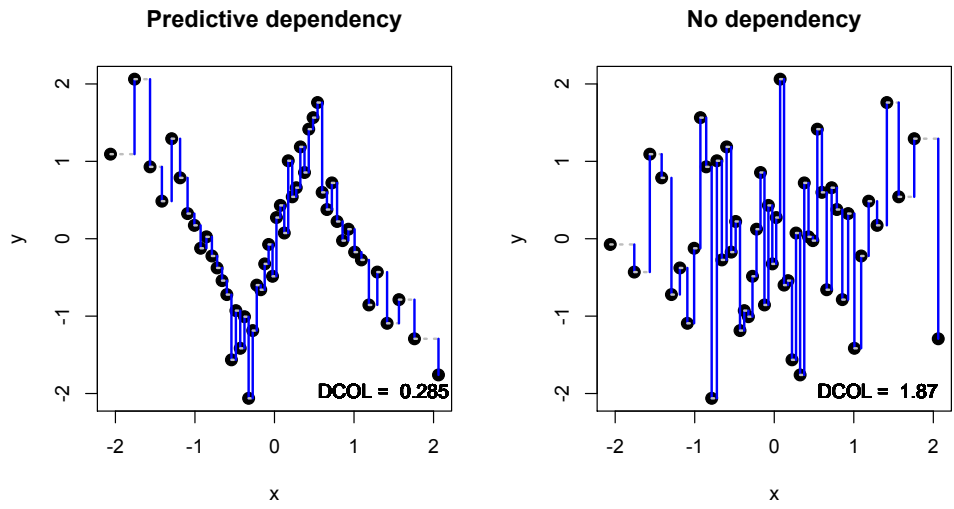

**Figure 1**. Illustration of DCOL. The DCOL score is calculated from the average squared length of the blue bars.

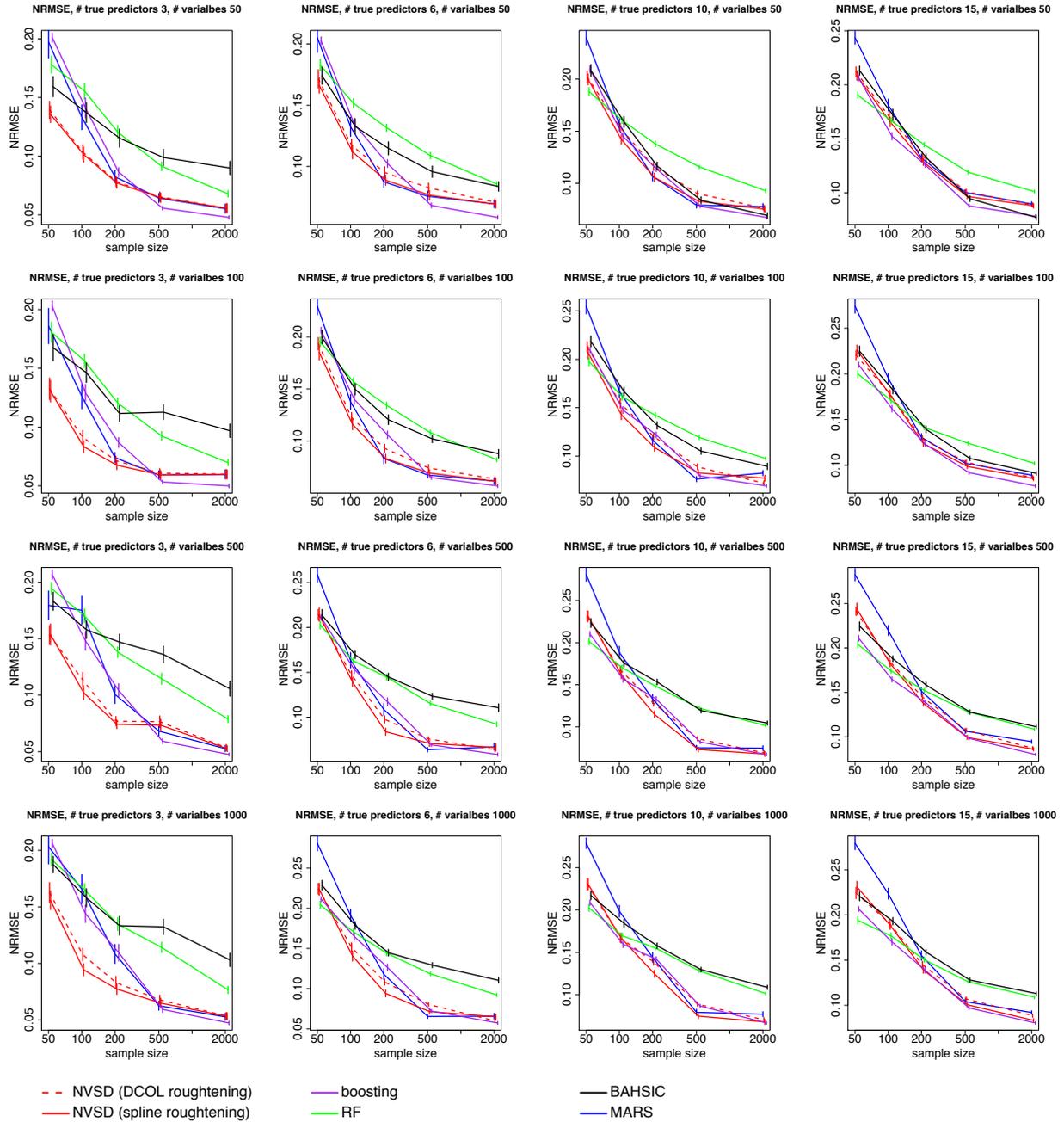

**Figure 2.** Simulation results. The average NRMSE are plotted against the sample size. Different sub-plots represent different true number of predictors (columns) and total number of variables (rows). The results were based on 50 simulations at each parameter setting. The ±standard error of the estimate is shown as a vertical bar.

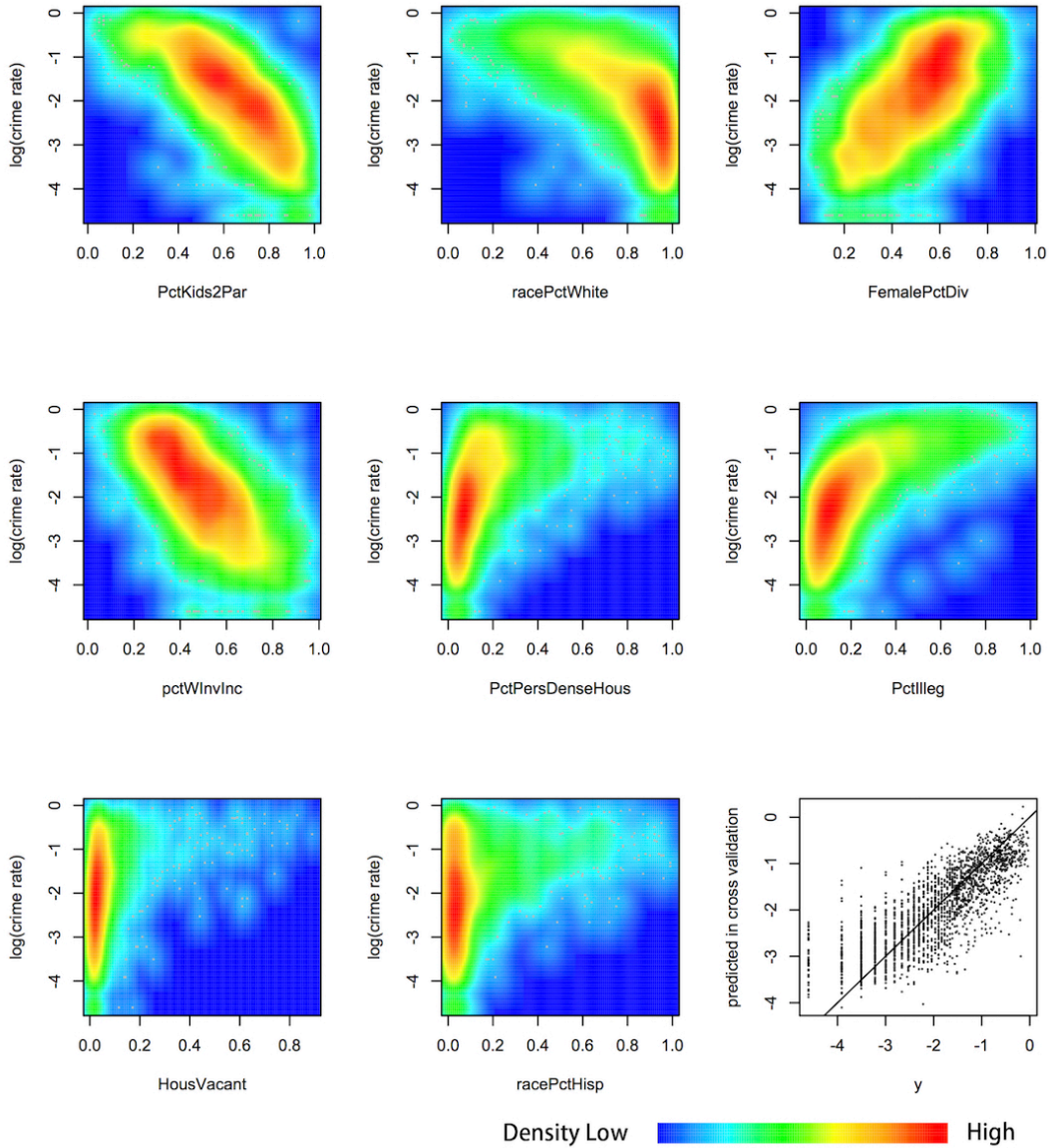

**Figure 3.** Variables selected for the crime rate data. The outcome variable is plotted against the selected variables one at a time using density scatter plots. The lower-right plot shows the scatterplot of predicted values in 5-fold cross-validation against the true values.

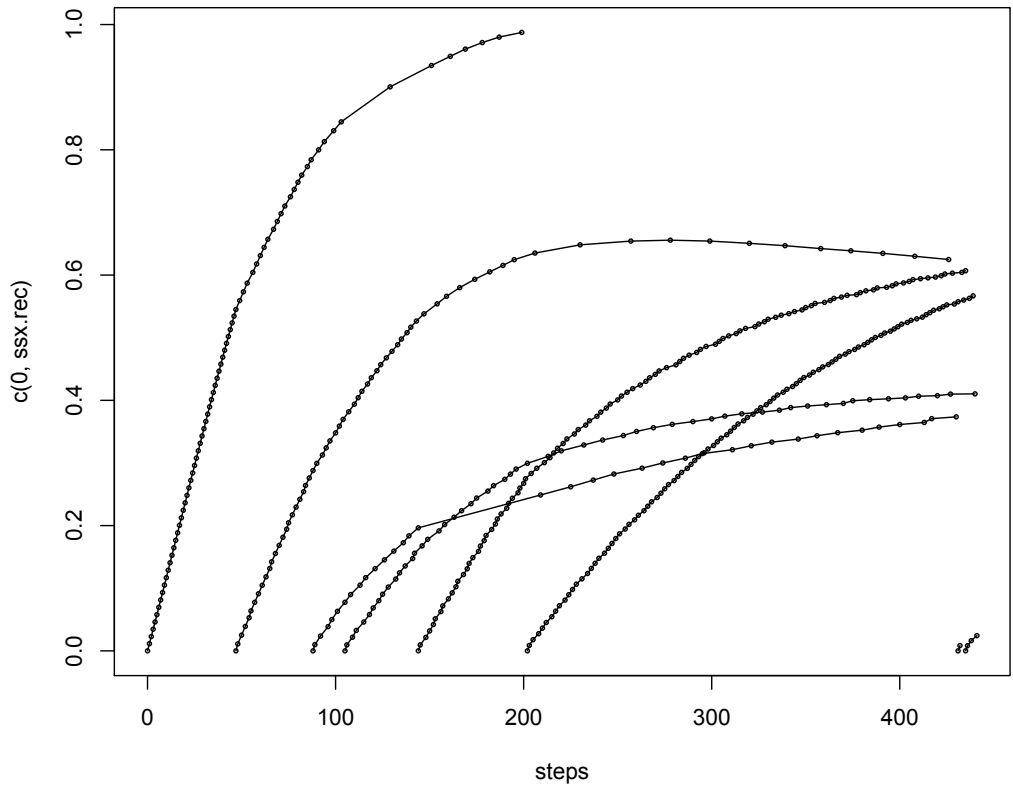

**Figure 4.** A example of the heuristic solution path plot generated from the crime rate data.

# Supplementary Materials for "Nonlinear variable selection with continuous outcome: a fully nonparametric incremental forward stagewise approach" by Tianwei Yu

## 1. Appendix A
**Eq. 1.**

We assume the relationship between $Y$ and $X$ is $Y = f(X) + \varepsilon$, where $f()$ is a continuous function on finite support, and with finite values and finite first derivative everywhere, and $\varepsilon$ is i.i.d. additive noise with mean 0 and variance $\sigma^2$.

Let the data points be ordered based on $x$ values: $(x_i, y_i): x_1 \leq x_2 \leq \ldots \leq x_n$, and $\Delta_i = y_{i+1} - y_i$, under the condition that $f()$ is on finite support, and has finite value and finite first derivative everywhere, we want to show that $s(\Delta) = \frac{1}{n-2}\sum_{i=1}^{n-1}(\Delta_i - \bar{\Delta})^2 \simeq 2\sigma^2$ when $n$ is large.

$$s(\Delta)$$
$$= \frac{1}{n-2}\sum_{i=1}^{n-1}(\Delta_i - \bar{\Delta})^2$$
$$= \frac{1}{n-2}\sum_{i=1}^{n-1}(\Delta_i^2 + \bar{\Delta}^2 - 2\bar{\Delta}\Delta_i)$$
$$= \frac{1}{n-2}\sum_{i=1}^{n-1}(\Delta_i^2) + \frac{1}{n-2}\sum_{i=1}^{n-1}(\bar{\Delta}^2) - \frac{2}{n-2}\sum_{i=1}^{n-1}(\bar{\Delta}\Delta_i)$$

Given $\bar{\Delta} = \frac{1}{n}(f(x_n) - f(x_1) + \varepsilon_n - \varepsilon_1)$, and $f()$ is finite, it is easy to show the latter two terms, $\frac{1}{n-2}\sum_{i=1}^{n-1}(\bar{\Delta}^2) - \frac{2}{n-2}\sum_{i=1}^{n-1}(\bar{\Delta}\Delta_i)$, approach zero as $n \to \infty$.

We next consider the first term,

$$\frac{1}{n-2}\sum_{i=1}^{n-1}(\Delta_i^2)$$
$$= \frac{1}{n-2}\sum_{i=1}^{n-1}(\Delta_i^2)$$
$$= \frac{1}{n-2}\sum_{i=1}^{n-1}(f(x_{i+1}) - f(x_i) + \varepsilon_{i+1} - \varepsilon_i)^2$$
$$= \frac{1}{n-2}\sum_{i=1}^{n-1}\Big(f^2(x_{i+1}) + f^2(x_i) - 2f(x_{i+1})f(x_i) + (\varepsilon_{i+1} - \varepsilon_i)^2$$
$$\quad + 2(\varepsilon_{i+1} - \varepsilon_i)(f(x_{i+1}) - f(x_i))\Big)$$

$$= \frac{1}{n-2}\sum_{i=1}^{n-1} f^2(x_{i+1}) + \frac{1}{n-2}\sum_{i=1}^{n-1} f^2(x_i) - \frac{2}{n-2}\sum_{i=1}^{n-1} f(x_{i+1})f(x_i)$$
$$+ \frac{1}{n-2}\sum_{i=1}^{n-1} (\varepsilon_{i+1} - \varepsilon_i)^2 + \frac{2}{n-2}\sum_{i=1}^{n-1} (\varepsilon_{i+1} - \varepsilon_i)(f(x_{i+1}) - f(x_i))$$

as $n \to \infty$, $x_{i+1} - x_i \to 0$, we can use first-order Taylor expansion,

$$\cong \frac{1}{n-2}\sum_{i=1}^{n-1} f^2(x_{i+1}) + \frac{1}{n-2}\sum_{i=1}^{n-1} f^2(x_i)$$
$$- \frac{2}{n-2}\sum_{i=1}^{n-1} (f(x_i) + f'(x_i)(x_{i+1} - x_i))f(x_i)$$
$$+ \frac{1}{n-2}\sum_{i=1}^{n-1} (\varepsilon_{i+1} - \varepsilon_i)^2 + \frac{2}{n-2}\sum_{i=1}^{n-1} (\varepsilon_{i+1} - \varepsilon_i)(f(x_{i+1}) - f(x_i))$$
$$= \frac{1}{n-2}\sum_{i=1}^{n-1} f^2(x_{i+1}) - \frac{1}{n-2}\sum_{i=1}^{n-1} f^2(x_i) - \frac{2}{n-2}\sum_{i=1}^{n-1} f(x_i)f'(x_i)(x_{i+1} - x_i)$$
$$+ \frac{1}{n-2}\sum_{i=1}^{n-1} (\varepsilon_{i+1} - \varepsilon_i)^2 + \frac{2}{n-2}\sum_{i=1}^{n-1} (\varepsilon_{i+1} - \varepsilon_i)(f(x_{i+1}) - f(x_i))$$

$$= \frac{f^2(x_n) - f^2(x_1)}{n-2} - \frac{2}{n-2}\sum_{i=1}^{n-1} f(x_i)f'(x_i)(x_{i+1} - x_i) + \frac{1}{n-2}\sum_{i=1}^{n-1} (\varepsilon_{i+1} - \varepsilon_i)^2$$
$$+ \frac{2}{n-2}\sum_{i=1}^{n-1} (\varepsilon_{i+1} - \varepsilon_i)(f(x_{i+1}) - f(x_i))$$

Given $f(x)$ and $f'(x)$ is finite everywhere, and $x_1, \ldots, x_n$ are ordered from the smallest to the largest, the first term clearly approaches zero as $n$ is large. For the second term, let $C$ be the largest $|f(x_i)f'(x_i)|$ among all $i$.

$$\left| \frac{2}{n-2}\sum_{i=1}^{n-1} f(x_i)f'(x_i)(x_{i+1} - x_i) \right| \le \frac{2C}{n-2}\sum_{i=1}^{n-1} (x_{i+1} - x_i) = \frac{2C}{n-2}(x_n - x_1)$$

This terms approaches zero given the function is on finite support.

For the last term,

$$\frac{2}{n-2}\sum_{i=1}^{n-1} (\varepsilon_{i+1} - \varepsilon_i)(f(x_{i+1}) - f(x_i))$$

$$= \frac{2}{n-2}\sum_{i=1}^{n-1} \varepsilon_{i+1}(f(x_{i+1}) - f(x_i)) - \frac{2}{n-2}\sum_{i=1}^{n-1} \varepsilon_i(f(x_{i+1}) - f(x_i))$$
$$= \frac{2}{n-2}\sum_{i=1}^{n-1} \varepsilon_{i+1}f(x_{i+1}) - \frac{2}{n-2}\sum_{i=1}^{n-1} \varepsilon_{i+1}f(x_i) - \frac{2}{n-2}\sum_{i=1}^{n-1} \varepsilon_i f(x_{i+1})$$
$$+ \frac{2}{n-2}\sum_{i=1}^{n-1} \varepsilon_i f(x_i)$$

The four terms are similar. Given $\varepsilon_i's$ are $i.i.d.$ from a distribution, and can be freely permuted, the ordering of $f(x_i)$ can be ignored. Now we consider the sample covariance term $\frac{2}{n-2}\sum_{i=1}^{n-1}\varepsilon_{i+1}(f(x_{i+1}) - \bar{f})$, where $\bar{f}$ is the sample average of the $f(x_{i+1}), i = 1, \ldots, n$. Given independence of the signal and noise, the term converges to zero as $n \to \infty$.

$$\frac{2}{n-2}\sum_{i=1}^{n-1}\varepsilon_{i+1}(f(x_{i+1}) - \bar{f})$$
$$= \frac{2}{n-2}\sum_{i=1}^{n-1}\varepsilon_{i+1}f(x_{i+1}) - \frac{2}{n-2}\bar{f}\sum_{i=1}^{n-1}\varepsilon_{i+1}$$

And $f()$ is finite, the second term converges to zero as $n \to \infty$. Thus $\frac{2}{n-2}\sum_{i=1}^{n-1}\varepsilon_{i+1}f(x_{i+1})$ converges to zero as well.

Thus we have as $n \to \infty$, $s(\Delta) \cong \frac{1}{n-2}\sum_{i=1}^{n-1}(\varepsilon_{i+1} - \varepsilon_i)^2$, which converges to $2\sigma^2$.

## 2. Appendix B

**General roughening with smoother**

We assume all functions linking X variables and Y are continuous on finite support, and with finite values and finite first derivative everywhere. Here we assume a local weighted average smoother is used for the general roughening procedure. With the increase of sample size, the bandwidth of the smoother will shrink towards zero.

$y_i = f(x_{1,i}) + g(x_{2,i}) + \varepsilon_i$, and

$y_i^{new} = y_i + \theta(y_i - \tilde{y}_i) = y_i + \theta\left(y_i - \sum_j w_j y_j\right)$

where $w_j's$ are positive weights between zero and one, and sum up to one.

$$y_i^{new}$$

$$= y_i + \theta\left(y_i - \sum_j w_j y_j\right)$$

$$= y_i + \theta\left(f(x_{1,i}) - \sum_j w_j f(x_{1,j})\right) + \theta\left(g(x_{2,i}) - \sum_j w_j g(x_{2,j})\right)$$
$$+ \theta\left(\varepsilon_i - \sum_j w_j \varepsilon_j\right)$$

$$= y_i + \theta\left(f(x_{1,i}) - \sum_j w_j f(x_{1,j})\right) + (\theta + w_i)g(x_{2,i}) - \theta\left(\sum_{j \neq i} w_j g(x_{2,j})\right)$$
$$+ \theta\left(\varepsilon_i - \sum_j w_j \varepsilon_j\right)$$

As *N* goes to infinity, the smoother window shrinks towards zero, thus in the second term, only the $x_{1,j}$'s that are extremely close to $x_{1,i}$ will receive non-zero weight. As $\sum_j w_j = 1$, the second term tends to zero as *N* goes large.

On the other hand, given that the ordering has no bearing on $x_2$'s, thus $x_{2,j}, j \neq i$ can be considered as random i.i.d. samples drawn from the probability density of $X_2$. Hence $\sum_{j \neq i} w_j g(x_{2,j})$ has a mean of zero, and variance of $\sum_{j \neq i} w_j^2 \varphi^2$, assuming the standard deviation of $g(X_2)$ is $\varphi$. Hence we can write,

$$y_i^{new} \cong y_i + (\theta + w_i)g(x_{2,i}) + \omega,$$

where $\omega$ is the noise term with mean 0 and variance of $\sum_{j \neq i} \theta^2 w_j^2 \varphi^2 + \left((1 - w_i)^2 + \sum_{j \neq i} w_j^2\right)\theta^2 \sigma^2$. Hence after a step of roughening, the relative contribution of the predictive variable that is not the basis of the current roughening step would be increased. The same argument can extend to later iterations of the roughening process, as well as the scenario of multiple predictive variables.

# 3. Appendix C
# Supplementary figures

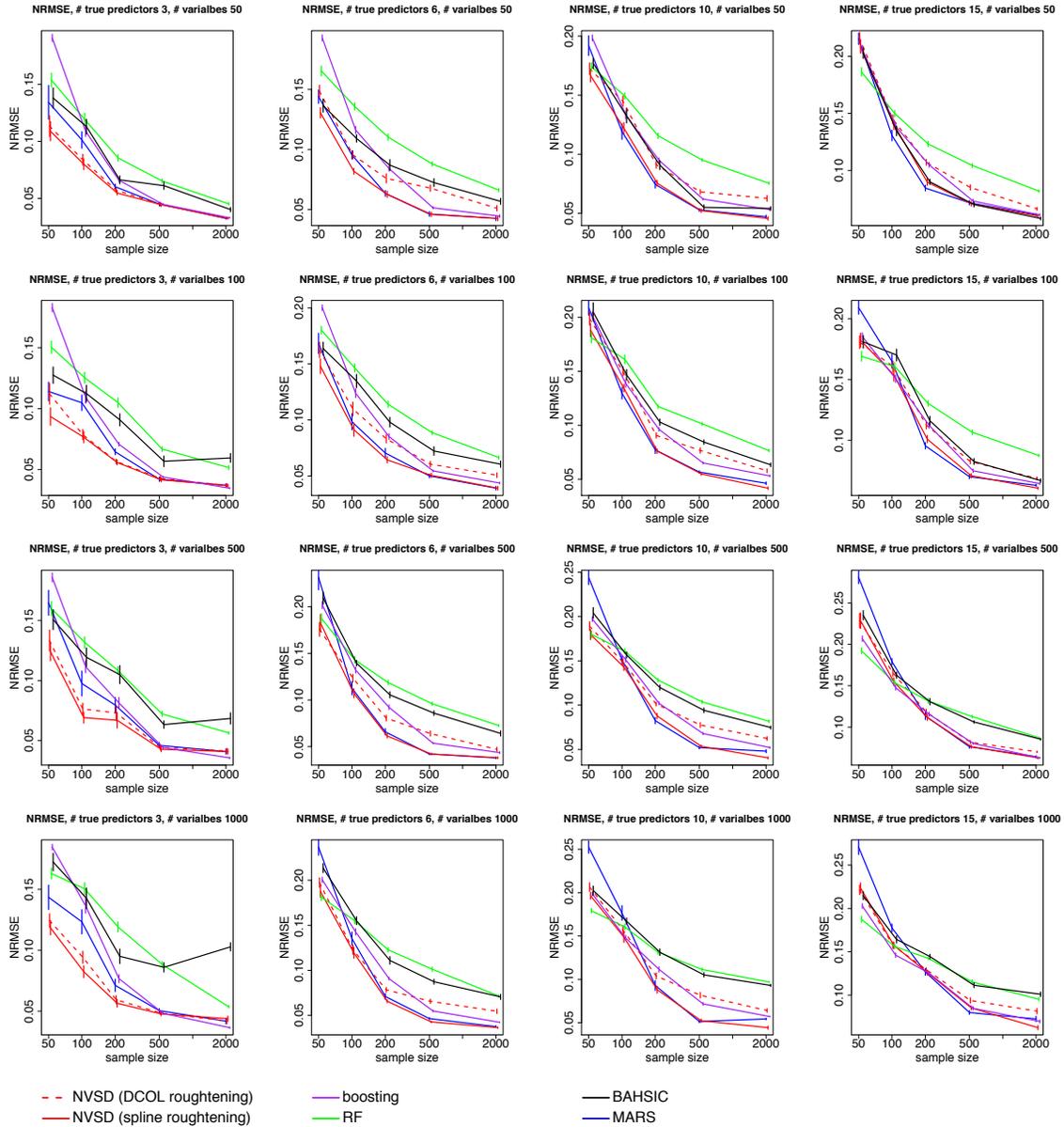

**Supplementary Figure 1.** Simulation NRMSE results using normally distributed X variables. The results were based on 50 simulations at each parameter setting. The ±standard error of the estimate is shown as a vertical bar.

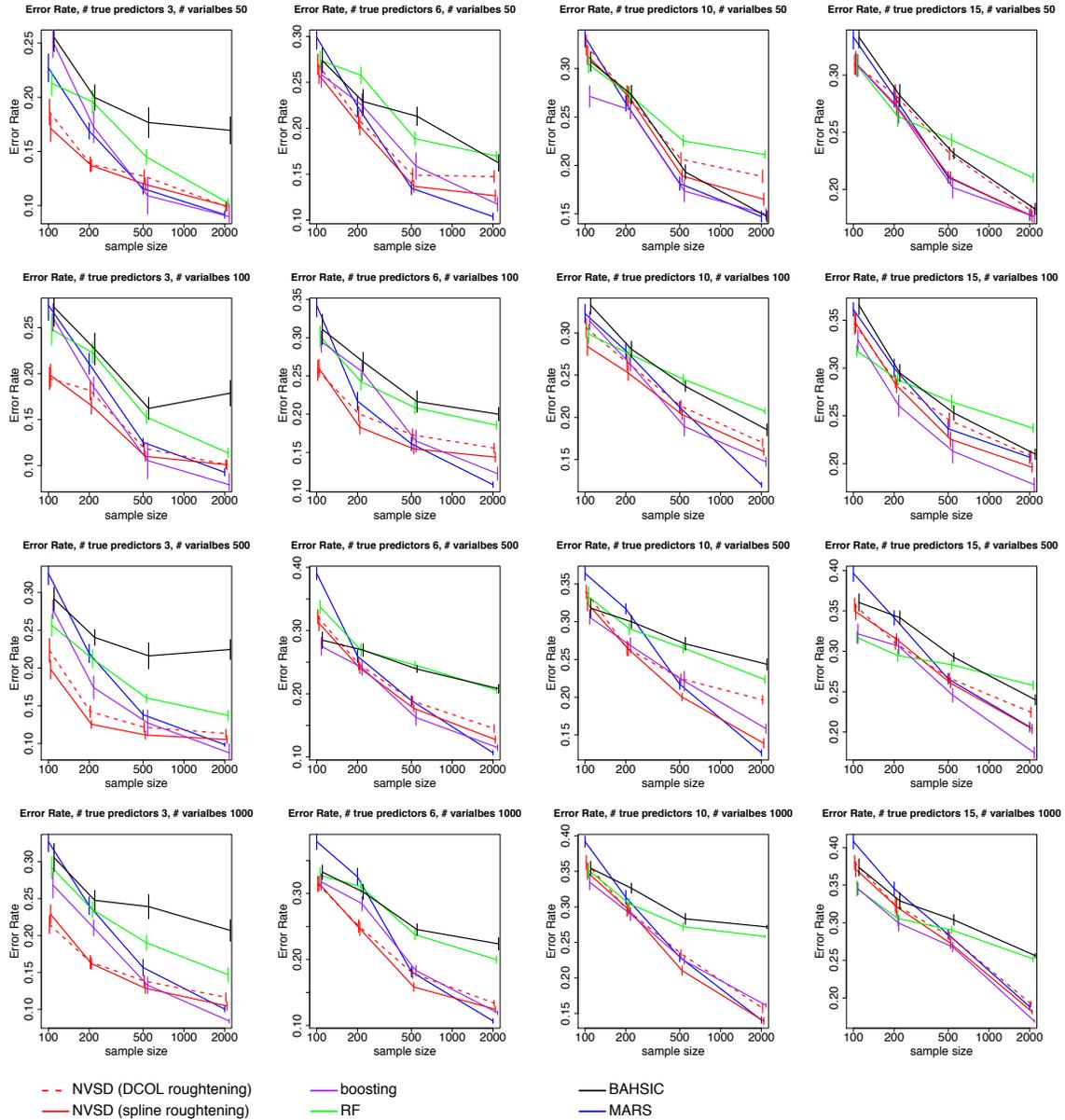

**Supplementary Figure 2.** Results from simulations with 0/1 outcome. The results were based on 50 simulations at each parameter setting. The ±standard error of the estimate is shown as a vertical bar.

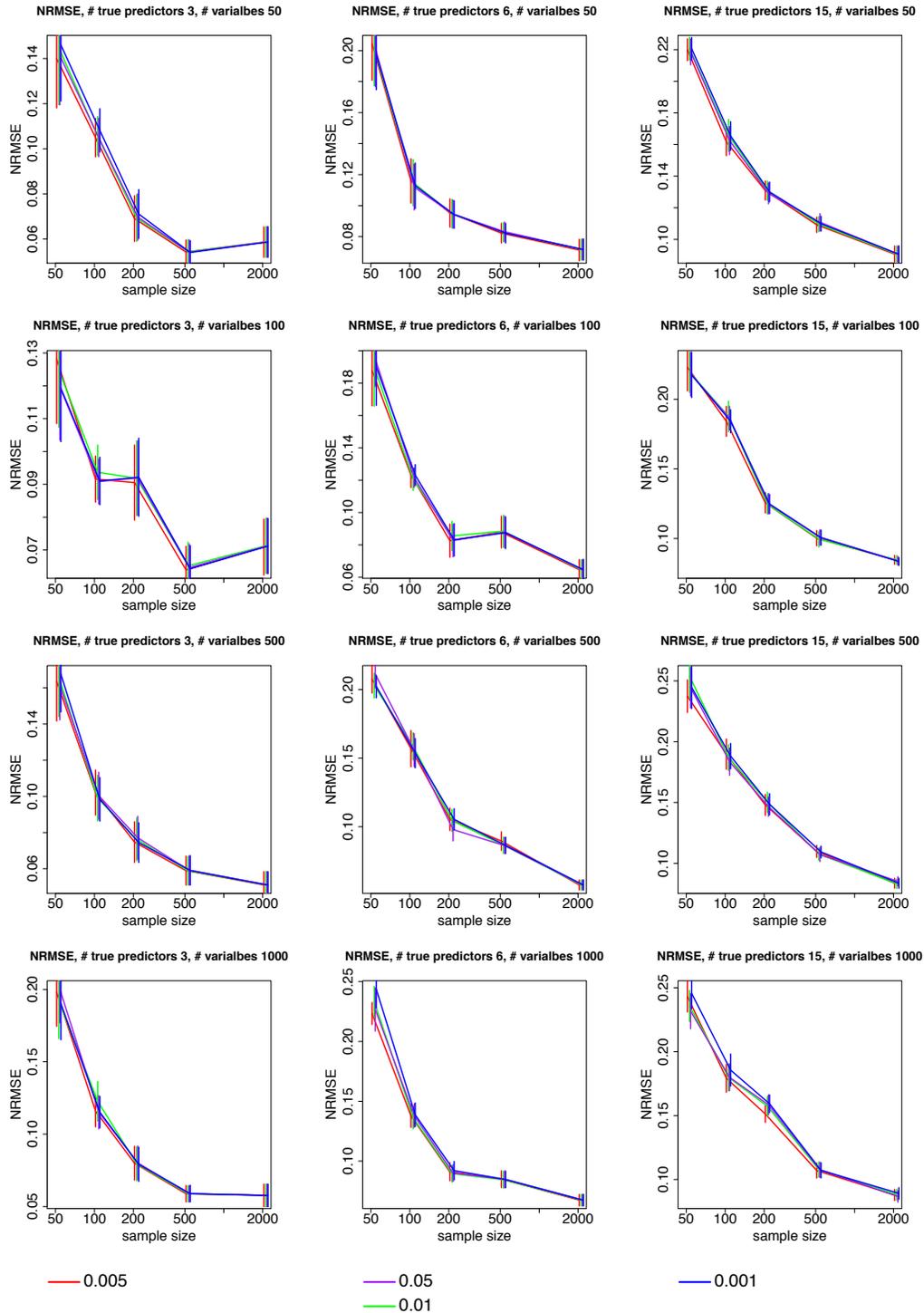

**Supplementary Figure 3.** Simulation NRMSE results using uniformly distributed X variables, DCOL roughening, and at different step sizes. The results were based on 10 simulations at each parameter setting. The ±standard error of the estimate is shown as a vertical bar.

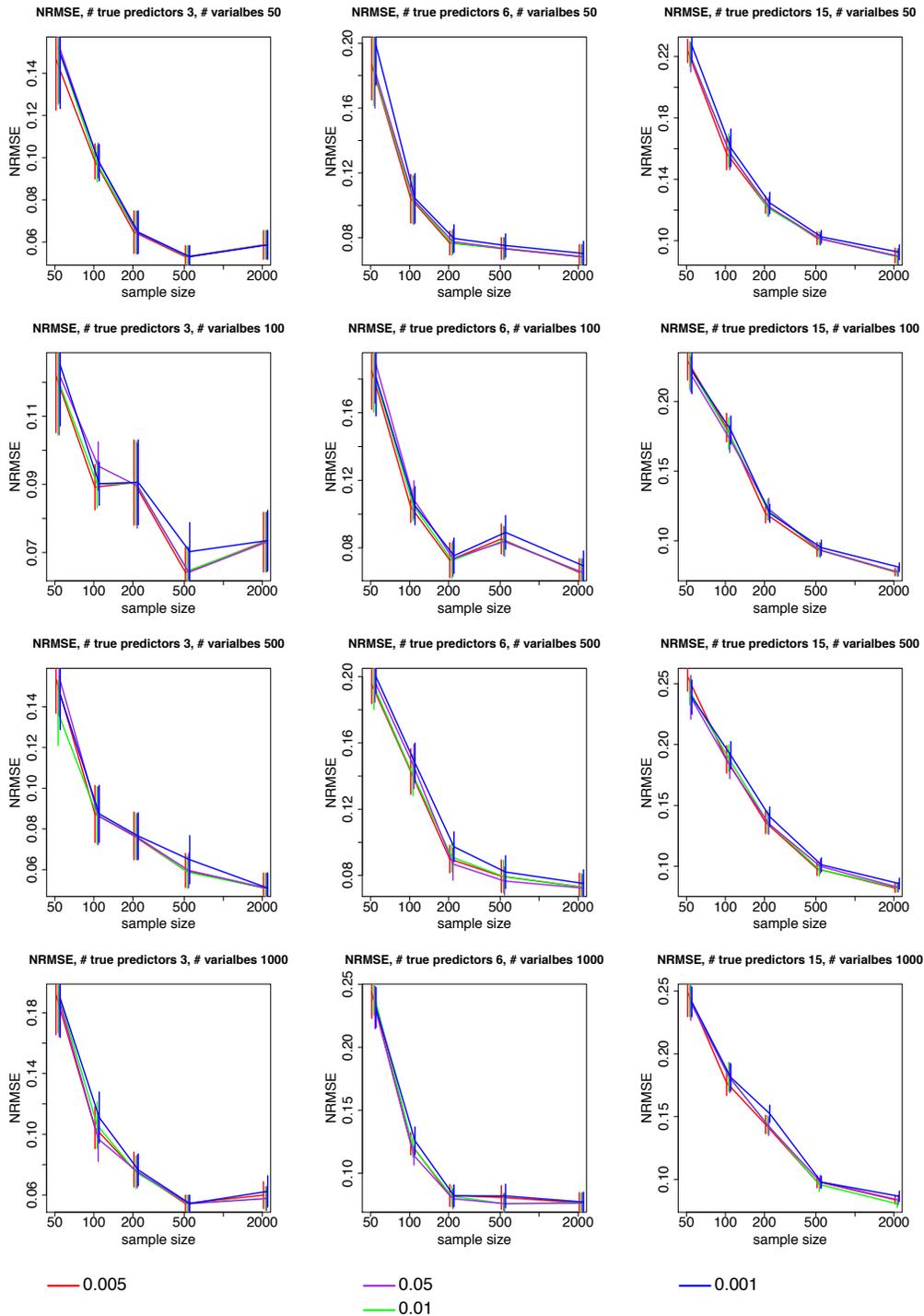

**Supplementary Figure 4.** Simulation NRMSE results using uniformly distributed X variables, spline roughening, and at different step sizes. The results were based on 10 simulations at each parameter setting. The ±standard error of the estimate is shown as a vertical bar.

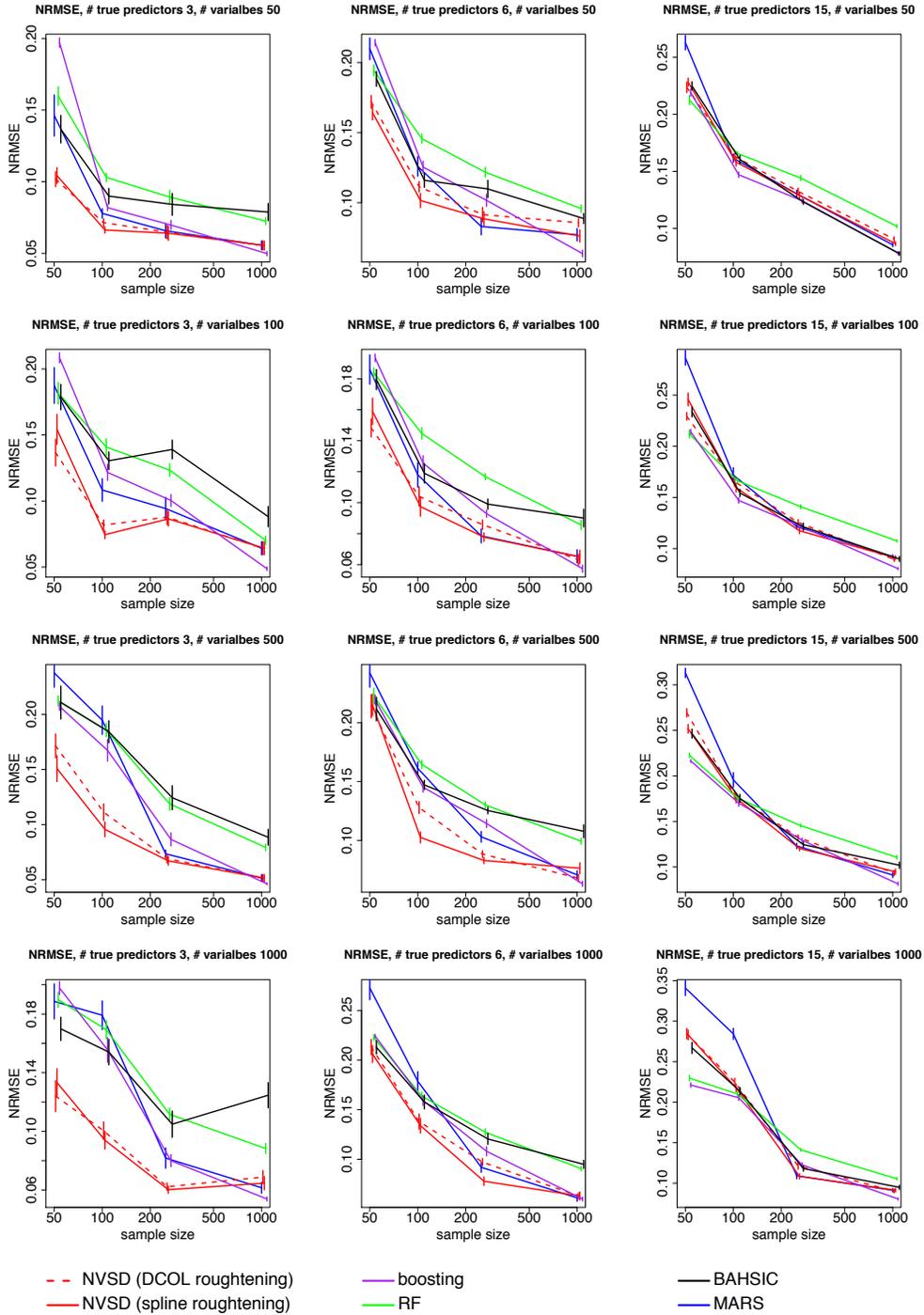

**Supplementary Figure 5.** Simulation NRMSE results using uniformly distributed X variables, and identity correlation matrix. The results were based on 50 simulations at each parameter setting. The ±standard error of the estimate is shown as a vertical bar.

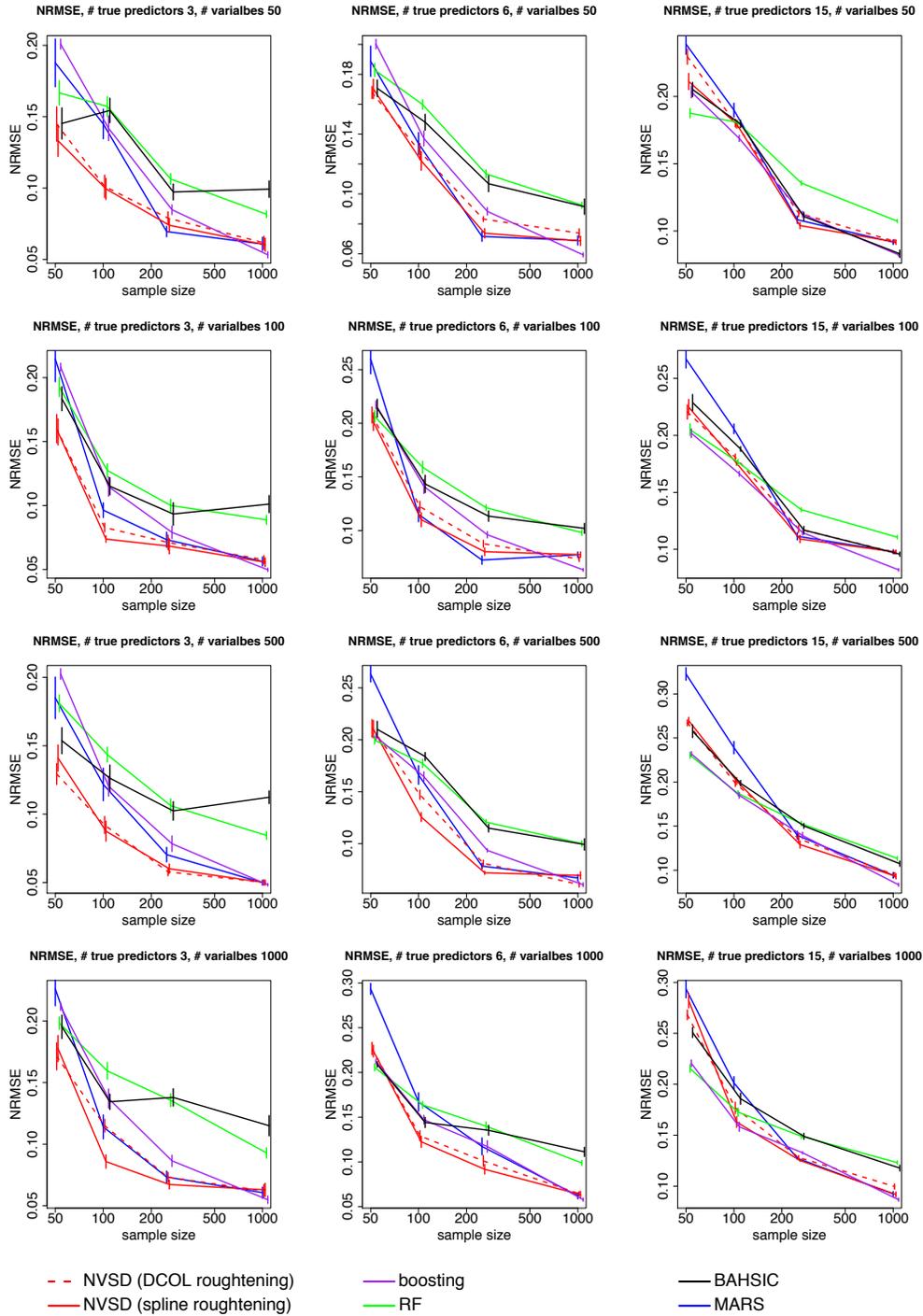

**Supplementary Figure 6.** Simulation NRMSE results using uniformly distributed X variables, and weaker correlation matrix - Consider the covariance matrix is $\Sigma$ from the real data. In the modified matrix $\Sigma'$, $\sigma'_{ij} = sign(\sigma_{ij})\sigma_{ij}^2$. The results were based on 50 simulations at each parameter setting. The ±standard error of the estimate is shown as a vertical bar.

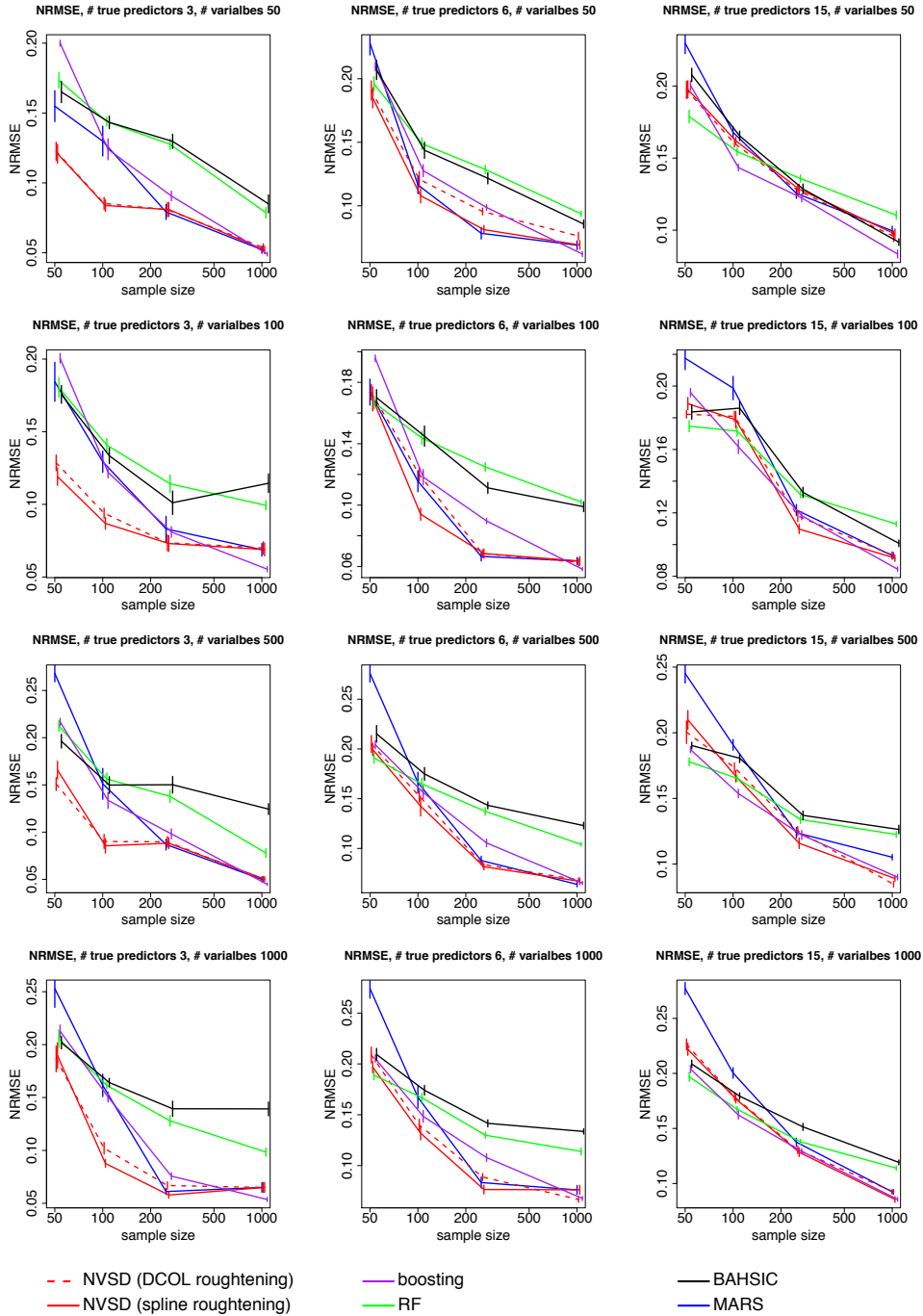

**Supplementary Figure 7.** Simulation NRMSE results using uniformly distributed X variables, and stronger correlation matrix - Consider the covariance matrix is $\Sigma$ from the real data. In the modified matrix $\Sigma'$, $\sigma'_{ij} = sign(\sigma_{ij})\sqrt{|\sigma_{ij}|}$. The results were based on 50 simulations at each parameter setting. The ±standard error of the estimate is shown as a vertical bar.

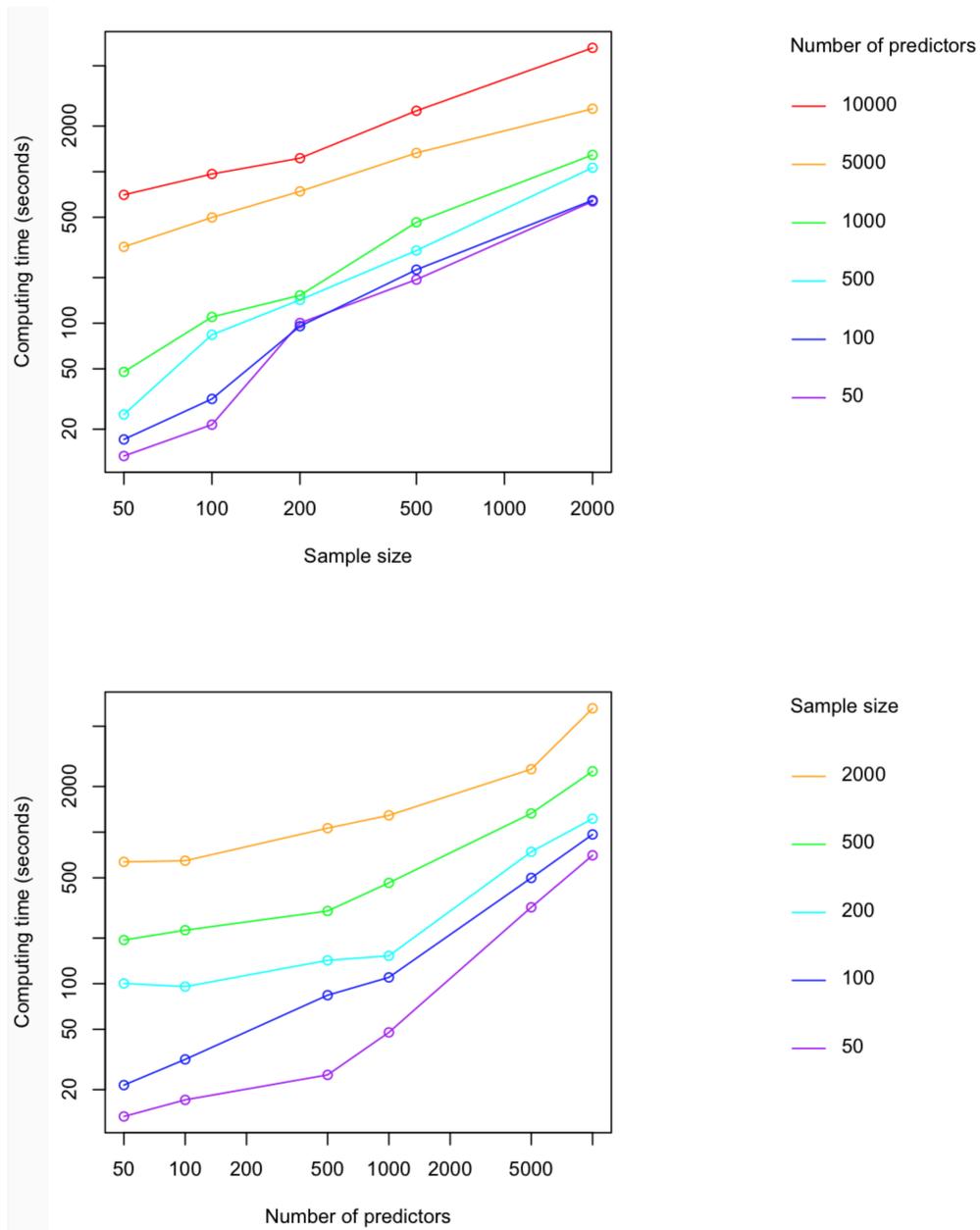

**Supplementary Figure 8.** The computing time on datasets. Each point was based on an average of 4 simulations. The figure was generated using step size of 0.01, stopping alpha of 0.01, and in the setting of 6 true predictors. The average computing time of DCOL roughening and spline roughening is reported in the figure.

# 4. Appendix D
# Selected genes (ENTREZ gene ID) for GSE10255 dataset

| Module 1 | 3133 29116 5996 2512 27089 8660 8407 6629 3281 596 84864 2038 6772 50486 440270 8418 3535 2944 283 2799 3229 54463 5744 10449 64760 64324 3352 79587 7182 6207 4594 2947 5217 3563 9616 59349 2827 54623 8139 5603 9783 8754 9725 222642 7431 59067 6509 79627 64089 11314 6689 7289 80131 64097 3269 10763 1396 79930 23099 590 1579 7347 55801 1456 4799 2057 4951 50618 8487 22 29969 1572 9427 7367 5617 2859 4700 83475 63933 26747 23385 10930 3621 26050 29767 9448 80201 1382 7064 6176 869 27349 6850 1297 11311 9718 8509 7881 79158 57092 7358 9143 64769 9265 1238 11193 9723 941 54145 56995 28755 8802 8433 9022 1136 3906 2828 27304 3670 998 8891 55722 28965 2151 4666 51755 51334 2560 908 9786 1974 9275 8824 9708 |
|---|---|
| Module 2 | 3134 1508 10379 5685 6515 2040 23590 25796 913 9334 3097 664 6890 54545 84934 4601 22902 51275 3398 64718 4783 54765 11000 4225 2918 10396 4092 22806 9140 8932 55854 10439 22843 10452 23263 221830 3606 1843 5601 23379 9895 6932 57001 54937 3480 26271 10902 2915 445 55640 22845 144404 51200 5031 8161 6301 11163 29072 85865 5607 93487 7249 26133 23048 6710 51760 56649 22989 10193 27306 5756 10286 654434 55026 51513 3688 100129250 51236 54221 57115 23390 10351 9391 9843 54718 84658 9808 100128553 22913 79772 64425 135948 6248 284541 9580 56255 23585 11086 65220 11024 79908 1437 10287 6443 26130 9254 683 55958 7124 389906 781 27165 51083 10127 7351 23568 23705 9542 55180 90864 |
| Module 3 | 3107 5142 5682 2353 3429 55329 9917 1200 713 79956 54881 41 2802 81567 2222 11041 2568 9351 10135 7317 1947 55133 51118 2049 9180 55128 23522 8819 11221 55781 28985 9342 9968 117246 128872 6854 3695 5481 8741 3357 2319 7096 29766 79003 9950 4929 11056 27247 5590 115653 3645 55716 123 351 8115 11257 7454 2788 2483 64061 57818 79614 5279 313 57062 56922 6604 27343 8864 3329 10954 4882 4811 8894 29108 9882 1828 6799 3589 27122 4850 57048 57212 432 8647 9931 10436 80069 11190 2953 80199 23466 11014 56548 8365 9173 1983 10795 22920 138151 8887 140545 55833 55658 64856 58504 84124 |
| Module 4 | 3106 51668 4942 2926 4809 3655 6749 7357 348 3433 4708 10412 2354 10221 26092 23015 6351 85352 23505 1654 23191 10724 3034 51072 6774 8867 3808 656 1550 23475 2946 64682 224 25813 323 83463 4163 64175 4627 6641 11007 23484 138046 825 51161 23042 54762 5994 5912 186 91355 9991 9938 146542 55603 60489 6323 647070 79986 56981 |
| Module 5 | 3157 10577 6635 3135 53339 64759 2791 28831 28908 4051 5475 6446 5054 26034 11119 29922 3934 4709 200576 3223 25780 10223 55423 63934 27293 10740 57369 56260 3885 7385 1749 57152 6605 6258 9454 |
| | 26091 326626 58189 39 11340 10020 8853 8226 54893 10913 26135 7465 55592 3249 56949 91353 55227 490 3276 7638 811 51078 23760 8823 23468 28990 3577 |
| Module 6 | 5791 3105 518 80148 11332 645 3925 55226 90861 6375 2051 3587 5250 8454 133 9844 4686 5732 23149 3434 23245 6525 58473 7538 3003 4329 91543 10970 10146 23240 10949 26521 25928 7299 1186 3725 7128 9702 2152 5800 55799 55751 573 55601 80135 1959 11092 712 23424 285830 440738 |
| Module 7 | 7905 7466 537 51002 5341 10866 915 51042 84981 4318 8804 5934 1234 392 6414 25836 824 9235 90141 308 5476 6261 3938 645784 3507 114049 10762 10370 65110 3815 2700 5515 5830 100129656 8609 969 7673 51283 11161 5730 51281 51752 64766 5896 8840 7534 4793 54938 58528 8631 55821 5003 9684 56107 9200 4111 4139 13 1613 55654 23677 8076 81892 |
| Module 8 | 23179 3033 6342 801 5478 115353 54919 634 23001 10617 56950 797 5473 9425 339562 7158 91782 486 63943 2324 29911 2230 57096 64837 784 6595 55576 10632 51103 3036 7001 2058 515 8820 9440 23244 1841 |
| Module 9 | 54585 9473 7846 6637 4616 4706 6448 10964 7334 221749 1465 9926 383 6185 2867 23518 29990 3514 55062 10464 54584 5407 2634 51306 24149 3706 11083 3726 5049 2633 605 54808 22868 83448 5627 3301 6653 63920 10501 302 7070 644096 1278 10430 8318 8792 9025 10045 7704 3237 27350 23589 10319 161291 57104 6642 10586 5499 4067 114884 4815 8870 57610 53905 4284 9348 |
| Module 10 | 8073 7805 9130 50807 100130100 26268 5292 4245 1230 1112 54855 4238 4192 25814 5873 6615 9583 11137 4595 8985 9114 26548 80145 7424 1397 51191 471 91304 1543 93643 30848 8915 2140 10039 8065 619426 27115 29887 5869 5276 55743 4084 79731 |
| Module 11 | 4282 55617 910 113791 4005 55565 10989 10962 51312 23643 22863 5094 80781 3437 1314 28232 23645 5428 5496 5373 597 6907 929 11328 10384 10001 8503 51099 10542 5865 79139 55340 11034 55288 5979 637 79882 1839 23429 321 8036 10190 414771 9196 |
| Module 12 | 2308 3336 6636 757 9759 222487 1514 2589 10776 6286 55303 4753 3101 65990 6903 57379 54477 4330 26105 11171 7050 28955 28958 1524 9737 3868 5910 |
| Module 13 | 5747 5686 29087 3065 51134 51274 10212 200734 26227 652493 5050 4086 5305 6427 27020 51257 9394 4057 10209 1633 301 54973 3201 5256 5698 1741 54619 55844 3566 4867 1808 |
| Module 14 | 4208 2078 2135 79631 79080 1289 23228 29803 317781 6726 1193 2730 2653 1236 7430 114932 10864 54920 6310 56914 84747 57731 51297 3126 1291 3021 57804 4617 |
| Module 15 | 240 10971 4869 5269 60481 902 3459 23586 1265 5774 51371 79621 1847 2194 79829 54927 10409 50865 80746 7415 6434 10021 51728 165 23649 |
| Module 16 | 2004 352961 3638 7402 55196 2766 150759 3430 10404 51478 4722 8566 54739 10561 26060 6352 22881 5360 3916 779 51149 2147 10969 |
| Module 17 | 972 3087 3843 3419 1479 4599 10133 64216 10367 2098 25804 5782 2907 55073 4317 1622 23392 967 1281 23363 6391 6741 |